\documentclass[letterpaper]{article} % DO NOT CHANGE THIS
\usepackage{aaai2026}  % DO NOT CHANGE THIS
\usepackage{times}  % DO NOT CHANGE THIS
\usepackage{helvet}  % DO NOT CHANGE THIS
\usepackage{courier}  % DO NOT CHANGE THIS
\usepackage[hyphens]{url}  % DO NOT CHANGE THIS
\usepackage{graphicx} % DO NOT CHANGE THIS
\usepackage{natbib}  % DO NOT CHANGE THIS AND DO NOT ADD ANY OPTIONS TO IT

\usepackage{pgfplots}
\usepackage{tikz}  
\pgfplotsset{compat=1.18}
\usepackage[T1]{fontenc}
\usepackage{graphicx}
\usepackage{booktabs}
\usepackage{multirow}
\usepackage{adjustbox} 
\usepackage{amsmath}
\usepackage{soul}
\usepackage{caption} % DO NOT CHANGE THIS AND DO NOT ADD ANY OPTIONS TO IT
\usepackage{algorithm}
\usepackage{algorithmic}

\usepackage{newfloat}
\usepackage{listings}
\DeclareCaptionStyle{ruled}{labelfont=normalfont,labelsep=colon,strut=off} % DO NOT CHANGE THIS
\floatstyle{ruled}
\newfloat{listing}{tb}{lst}{}
\floatname{listing}{Listing}
\usepackage{bibentry}
\title{On the Limitations of Rank-One Model Editing in Answering Multi-hop Questions}
\author {
    Zhiyuan He,
    Binghan Chen,
    Tianxiang Xiong,
    Ziyang Sun,
    Mozhao Zhu,
    Xi Chen
}
\affiliations {
    University College London\\
    \{zcabebx, binghan.chen.24, zcabtxi, zcabzsu, zcabmz0, zcabhej\}@ucl.ac.uk
}

\begin{document}

\maketitle

\begin{abstract}
% Recent advances in Knowledge Editing (KE), such as Rank-One Model Editing (ROME), have demonstrated the ability to efficiently update factual knowledge in transformer models while outperforming traditional methods like fine-tuning and in-context learning on single-hop fact updates. However, these methods exhibit significant limitations when applied to multi-hop reasoning tasks, where answering a question requires chaining multiple pieces of knowledge.
% In this work, we identify and analyze three primary failure modes of ROME in multi-hop settings. First, we observe the "hopping-too-late" problem, where confining edits to a single layer disrupts the model’s reasoning chain, as later layers lack access to necessary intermediate representations. Second, the generalisation ability of ROME drops sharply when editing deeper layers. Third, we find that edited model sometimes overfits to edited knowledge, causing the model to incorrectly output the answer for edited hop regardless of other hops.

Recent advances in Knowledge Editing (KE), particularly Rank-One Model Editing (ROME), show superior efficiency over fine-tuning and in-context learning for updating single-hop facts in transformers. However, these methods face significant challenges when applied to multi-hop reasoning tasks requiring knowledge chaining.
In this work, we study the effect of editing knowledge with ROME on different layer depths and identify three key failure modes. First, the "hopping-too-late" problem occurs as later layers lack access to necessary intermediate representations. Second, generalization ability deteriorates sharply when editing later layers. Third, the model overfits to edited knowledge, incorrectly prioritizing edited-hop answers regardless of context.
To mitigate the issues of "hopping-too-late" and generalisation decay, we propose \textbf{Redundant Editing}, a simple yet effective strategy that enhances multi-hop reasoning. Our experiments demonstrate that this approach can improve accuracy on 2-hop questions by at least \textbf{15.5} percentage points, representing a \textbf{96\%} increase over the previous single-edit strategy, while trading off some specificity and language naturalness (Figure~\ref{fig:trade-off}).
%\footnote{Codes, datasets and results are available at \url{https://github.com/nickhezy/KE4MHQ}}

\end{abstract}

\section{Introduction}

In recent years, transformer-based \citep{vaswani2017attention} large language models (LLMs) have been widely adopted across various domains. As real-world facts evolve, efficiently and accurately updating stored knowledge has emerged as a critical research challenge \citep{DBLP:journals/corr/abs-2110-03215,mousavi2024outdate}. With modern LLMs growing increasingly large, traditional fine-tuning has become prohibitively expensive and challenging \citep{parthasarathy2024ultimate,betley2025emergent}. 
This demand has resulted in growth of Knowledge Editing (KE), in which Rank-One Model Editing (ROME) \citep{meng2023rome} is a representative technique. It can inject single-fact knowledge by updating the weights of one single MLP layer, outperforming other popular methods like fine-tuning and in-context learning \citep{meng2023rome, hase2023does}.

\begin{figure}[t]
\centering
\begin{tikzpicture}[scale=0.95]
\begin{axis}[
    every pin/.style={font=\tiny},
    xlabel={Language Score (Higher = Better)},
    ylabel={MHQ Accuracy (\%)},
    xmin=30, xmax=100,
    ymin=15, ymax=35,
    xtick={30,40,50,60,70,80,90,100},
    ytick={15,20,25,30,35},
    legend pos=south west,
    legend style={font=\small},
    grid=major,
    width=0.5\textwidth,
    height=0.4\textwidth,
    mark options={solid}
]

% Redundant-editing data points (blue circles)
\addplot[only marks, mark=*, mark size=2pt, red] coordinates {
    (82.1,18.2)
    (75.6,21)
    (64,21.6)
    (38,31.6)
};
\addlegendentry{Redundant Editing}

% ROME data points (red squares) - COLOR FIXED HERE
\addplot[only marks, mark=square*, mark size=2pt, blue] coordinates {
    (90.5,17)
};
\addlegendentry{ROME}

% Label points with edit layers
\node[pin=90:{$\mathsf{[5]}$}] at (axis cs:90.5,17) {};
\node[pin=90:{$\mathsf{[5,20]}$}] at (axis cs:82.1,18.2) {};
\node[pin=90:{$\mathsf{[5,10,20]}$}] at (axis cs:75.6,21) {};
\node[pin={[pin distance=8mm]90:{$\mathsf{[5,10,15,20]}$}}] at (axis cs:64,21.6) {};
\node[pin=10:{$\mathsf{[5,8,11,15,17,20]}$}] at (axis cs:38,31.6) {};

% Pareto frontier line
\addplot[black, dashed, thick] coordinates {
    (90.5,17)
    (82.1,18.2)
    (75.6,21)
    (64,21.6)
    (38,31.6)
};

\end{axis}
\end{tikzpicture}
\caption{Trade-off between MQuAKE multi-hop question answering accuracy and language score on COUNTERFACT single-hop questions. \textcolor{blue}{Square marker} denotes the original ROME editing with layer selected by causal tracing, while \textcolor{red}{circles} show redundant-editing configurations with layer combinations in brackets.}
\label{fig:trade-off}
\end{figure}
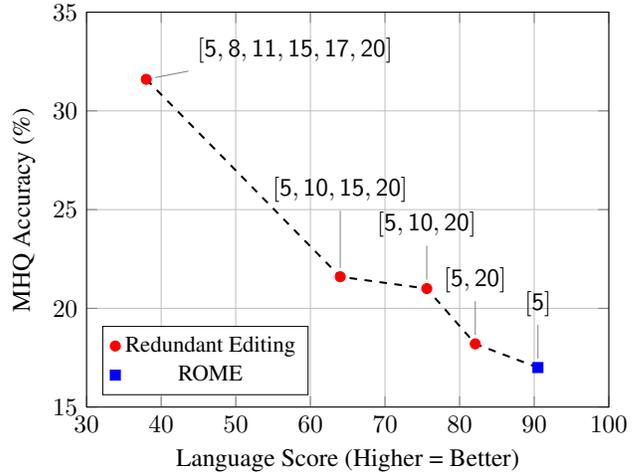

ROME works well on single-step knowledge tasks but struggles with complex questions like multi-hop reasoning \citep{DBLP:conf/emnlp/ZhongWMPC23}. \citet{DBLP:journals/corr/hopping-too-late} demonstrate that successful multi-hop reasoning depends critically on the relative layer positions where hop knowledge is stored. Given \citet{hase2023does,liu2025unlocking}'s finding that layer depth minimally impacts ROME's editing efficacy, we investigate how inserting knowledge at varying layer depths affects multi-hop reasoning.

Our research shows that ROME has three significant shortcomings in multi-hop tasks: (1) The "hopping-too-late problem" \citep{DBLP:journals/corr/hopping-too-late} occurs when hop-2 knowledge is stored in earlier layers than hop-1 knowledge, breaking the model’s internal reasoning chain. (2) Generalization capability drops rapidly when editing deeper layers, making edits more sensitive to question phrasing. (3) Overfit to edited knowledge regardless of the context question.

\begin{figure}[t]
    \centering
    \includegraphics[width=\linewidth]{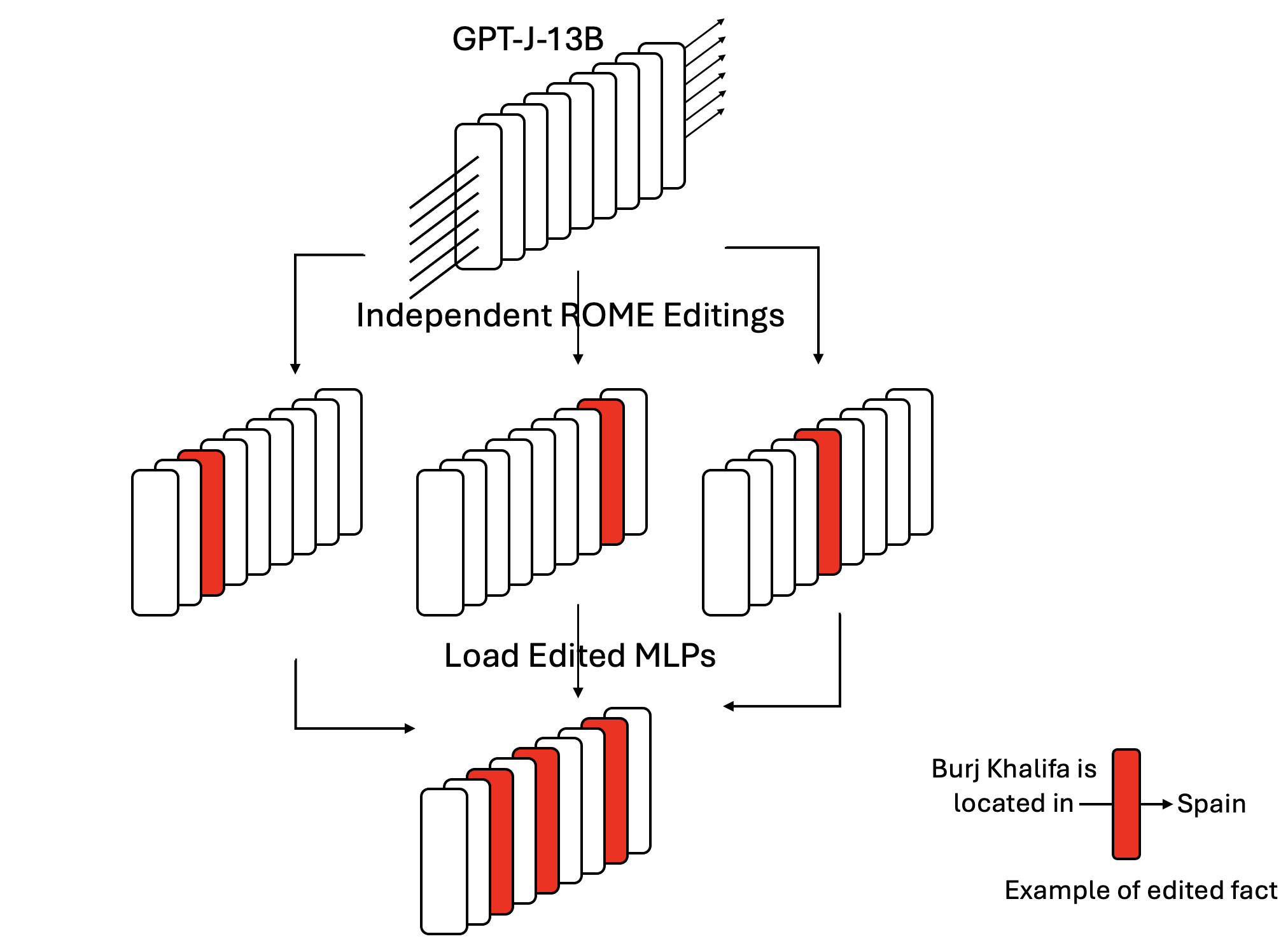} % Adjust width as needed
    \caption{Redundant Editing strategy: insert copies of a same knowledge into multiple layers.}
    \label{fig:procedure}
\end{figure}

To address the two problems, we propose \textbf{Redundant Editing}, which injects the same knowledge into several MLP layers with different depths, as illustrated in Figure~\ref{fig:procedure}. On the MQuAKE \citep{DBLP:conf/emnlp/ZhongWMPC23} 2-hop questions (2HQ) dataset, our strategy improves multi-hop question accuracy by 15.5 percentage points, a 96\% increase over single-layer editing, while trading off some specificity and naturalness. 
We investigate why ROME reduces specificity and naturalness, demonstrating that its overly strong edited knowledge signal suppresses other critical information in hidden representations.
We analyze the trade-off between multi-hop reasoning ability and language scores (Figure~\ref{fig:trade-off}), providing practitioners with guidance for selecting a suitable amount of layers to edit based on task requirements.

In summary our work contributes in (1) Revealed ROME’s limitations and analyzed three key failure patterns: ``hopping-too-late'', generalisation decay and specificity loss. (2) Proposed and validated ``Redundant Editing'', achieving significant performance gains on multi-hop questions. (3) Analyzed the trade-offs between the multi-hop reasoning ability and language metrics like specificity and naturalness.

\section{Related Work}

One prominent approach in KE involves modifying the down-projection layers of the feedforward network modules within transformer architectures. Methods such as ROME and Mass-Editing Memory in Transformers (MEMIT) \citep{meng2023rome} exemplify this strategy. ROME enables efficient updates to factual knowledge by directly altering specific model weights, while MEMIT extends this capability to facilitate large-scale edits across multiple facts simultaneously.

Despite their innovative designs, these methods have raised concerns regarding their practical applicability. ~\citet{DBLP:conf/emnlp/Yang0TMSYS24,gupta2024rebuilding} observed that ROME could destabilize LLMs with as little as a single edit, leading to model collapse. Similarly, ~\citet{DBLP:conf/acl/GuptaRA24} demonstrated that scaling edits using ROME and MEMIT results in both gradual and catastrophic forgetting, where the model loses previously acquired knowledge and its ability to perform downstream tasks. Furthermore, \citet{thibodeau2022but} highlighted limitations in ROME's generalization capabilities, noting that edits often fail to propagate bidirectionally and may not generalize across synonymous terms, indicating a token-level rather than concept-level modification.

Multi-hop question answering (MHQ) serves as a critical benchmark for evaluating the reasoning abilities of LLMs. \citet{DBLP:journals/corr/hopping-too-late} found that LLMs resolve intermediate entities in early layers and complete subsequent reasoning in later layers. This layered processing suggests that confining edits to a single layer may disrupt the model's reasoning chain, leading to the "hop-too-late" problem, where later layers lack access to necessary intermediate representations. \citet{DBLP:conf/emnlp/ZhongWMPC23} introduced MQuAKE, a benchmark designed to assess whether edited models can correctly answer multi-hop questions that depend on updated facts. Their findings indicate that while current KE approaches can recall edited facts accurately, they often fail on multi-hop questions requiring reasoning over multiple pieces of information. To address these challenges, \citet{DBLP:journals/corr/abs-2410-06331} proposed IFMET, a novel locate-then-edit KE approach designed to edit both shallow and deep MLP layers. By incorporating multi-hop editing prompts and supplementary datasets, IFMET aims to locate and modify knowledge across different stages of reasoning, thereby improving performance on multi-hop factual recall tasks.

\section{Preliminary}

\subsection{Notations}

We follow \citet{meng2023rome} and represent each fact as a triple $(s, r, o)$, where $s$ is the subject, $r$ the relation, and $o$ the object. For each fact editing, we aim to learn a new triple $(s, r, o^*)$ with old one replaced. In this work, we focus on \textbf{two-hop questions} (2HQ), where the answer requires chaining two such fact tripples: e.g., to answer “Which country is the tallest building in the world located in?”, one must infer $(\texttt{TallestBuilding}, \texttt{Is}, \texttt{BurjKhalifa})$ and then $(\texttt{BurjKhalifa}, \texttt{LocatedIn}, \texttt{UAE})$.

% \subsection{Rank-One Model Editing}

% ROME~\cite{meng2023rome} updates model weights to insert factual edits via a closed-form solution. The weight modification $\Delta W$ for the MLP projection matrix is computed via least-squares optimization:
% \[
% \Delta W = (k_{\text{edit}}^{\top} k_{\text{edit}})^{-1} k_{\text{edit}}^{\top} (v_{\text{new}} - W k_{\text{edit}})
% \]
% where $k_{\text{edit}}$ is the activation key of the edited subject, $v_{\text{new}}$ is the desired new value vector, and $W$ is the original weight. This rank-one update injects the new fact while aiming to preserve model behavior elsewhere.

\subsection{Rank-One Model Editing}

ROME~\cite{meng2023rome} computes the minimum-norm weight update down-projection matrix $\Delta W$ that satisfies $(W + \Delta W)k_s = v_{o^*}$ while minimizing interference via least-squares:

\[
\Delta W = {(k_s^\top k_s)^{-1}k_s^\top}(v_{o^*} - Wk_s)
\]

where $k_s$ is the subject's input activation and $v_{o^*}$ is the desired output representation for the new object, both extracted from the model's forward passes (averaged across contexts). The rank-one update modifies $W$ to map $k_s \rightarrow v_{o^*}$ while minimizing interference with other inputs.

\begin{figure*}[ht]
    \centering
    \includegraphics[width=0.9\linewidth]{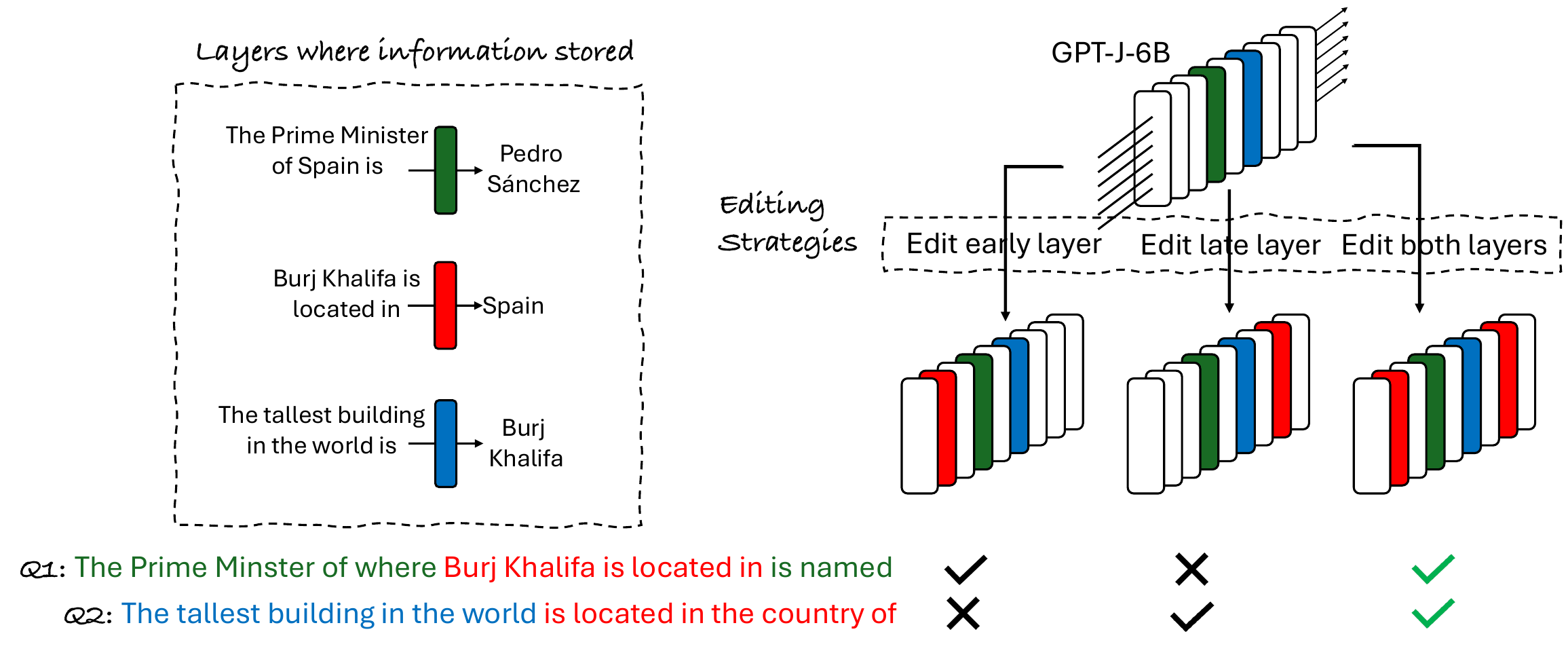} % Adjust width as needed
    \caption{Different multi-hop questions require the knowledge to be stored in different layers. Redundant insertions cover more multi-hop questions at the test time. This example is for illustration only, where correct hopping order does not always guarantee the correctness of the answer.
    % \footnote{This example is for illustration only, where correct hopping order does not always guarantee the correctness of the answer.
    }
    \label{fig:layers-view}
\end{figure*}

\section{Redundant Editing Strategy}

To overcome the challenges of solving multi-hop reasoning tasks in KE, we proposed Redundant Editing strategy — a methodology that inserts the same knowledge into multiple MLP layers simultaneously. As illustrated in Figure~\ref{fig:layers-view},  different 2HQ require the knowledge to be stored in different layers, for example, ``the tallest building in the world is Burj Khalifa'' has to be stored \textbf{in an earlier layer} than ``Burj Khalifa is located in Spain'' in order to build a valid internal reasoning chain for 2-hop question ``where is the tallest building in the world located?''. This is hard to achieve with only one single-layer knowledge injection. Inspired by this, our approach mitigates the “hopping-too-late” problem through injecting the same knowledge into multiple MLP layers with different depth.

Our methodology extends ROME by editing knowledge into multiple layers ranging from 5th to 20th.
To make sure each ROME edit is successful and learns complete features about the fact, we firstly execute ROME to different layers independently and then load the edited MLP layers to the original model, as illustrated in Figure~\ref{fig:procedure}. 
Through this Redundant Editing strategy, we make sure the knowledge to edit is stored in multiple copies in multiple layers so that when tested on multi-hop questions any of these copies can be used to build a reasoning chain.

% , ensuring comprehensive feature capture. We then integrate the individually edited MLP layers back into the original model (see Figure \ref{fig:procedure}. ). This "Redundant Editing" strategy stores multiple copies of the edited knowledge across layers, facilitating reliable reasoning chains for MHQ.

% this multi-layer strategy balances specificity and generalization capabilities, which makes the model maintain factual consistency across various formulations. We implement this on top of the original ROME framework using a modified \texttt{test\_multi\_rome()} pipeline. The model becomes more robust in handling hop orders and phrasing variations, which improves the multi-hop reasoning accuracy significantly while introducing a controlled trade-off in language fluency.

\begin{table*}[t]
\centering
\begin{tabular}{p{\textwidth}}
\toprule
Original MHQ: Which country is the tallest building in the world located in? [UAE] \\
\midrule
Edit \textbf{hop1} fact with ROME: The tallest building in the world is \st{Burj Khalifa} \textbf{Eiffel Tower} \\

Hop1 question (test for generalisability): Which building is the tallest in the world? [Eiffel Tower] \\

Hop2 question (test for specificity): Which country is the Eiffel Tower located in? [France] \\

2-hop question (test for gen., spec. and multi-hop chaining): Which country is the tallest building in the world located in? [France] \\
\midrule

Edit \textbf{hop2} fact with ROME : Burj Khalifa is located in \st{UAE} \textbf{Spain} \\

Hop1 question (test for specificity): Which building is the tallest in the world? [Burj Khalifa] \\

Hop2 question (test for generalisability): Which country is the Burj Khalifa located in? [Spain] \\

2-hop question (test for gen., spec. and multi-hop chaining): Which country is the tallest building in the world located in? [Spain] \\
\bottomrule
\end{tabular}
\caption{Examples of the fact edited and question tested on when the edited fact is hop1 and hop2 respectively.}
\label{tab:main-example}
\end{table*}

\section{Experiments}

We selected MQuAKE for its diverse multi-hop questions with explicit hop-level sub-questions and answers, which enable fine-grained reasoning analysis. The COUNTERFACT dataset provides complementary naturalness evaluations through specificity, fluency, and consistency metrics, addressing aspects beyond factual accuracy.

\subsection{MQuAKE Experiment Setup}

% We evaluated model editing performance on the GPT-J-6B model \cite{gpt-j} using the MQuAKE benchmark. Experiments were conducted on an NVIDIA RTX A6000 (48GB) GPU, with hyperparameters matching the original ROME configuration \cite{meng2023rome}. We experiment various strategies of editing different layers and make different numbers of Redundant Editing. 

% We evaluated model editing performance on the GPT-J-6B model \cite{gpt-j} using the MQuAKE benchmark. Experiments were conducted on an NVIDIA RTX A6000 (48GB) GPU, with hyperparameters (e.g., learning rate, batch size) matching the original ROME configuration \cite{meng2023rome}. We experimented with various strategies for editing different layers and applying different numbers of redundant edits.

We evaluated model editing performance on the GPT-J-6B (\citep{gpt-j}) model using the MQuAKE benchmark, with various strategies of editing different layers and make different numbers of Redundant Editing. 

% The primary metric for this evaluation was the accuracy of the model edit – specifically, whether the predicted tokens match any of the provided correct answers for the multi-hop questions. The results are shown in the table ~\ref{tab:main}, where each row indicates the MLP layer(s) in which multi-ROME is applied. 
We evaluate on a curated subset of the MQuAKE dataset, focusing on two testing scenarios: (1) edited knowledge is used in the \emph{first} hop of a 2HQ (240 instances), (2) edited knowledge is used in the \emph{second} hop of a 2HQ (359 instances).
For each scenario, we care about 3 types of question answering accuracies:
\begin{itemize}
    \item \textbf{Edited hop accuracy} It assesses how well the knowledge editing is generalized to a rephrased prompt querying for the knowledge edited.
    \item \textbf{Unedited hop accuracy} It assesses if the knowledge editing is specific enough to leave the other unedited knowledge unchanged.
    \item \textbf{2-hop question accuracy} It assesses if the edited knowledge can be used for a multi-step reasoning, which is closer to real-world LLM applications.
\end{itemize}

Table~\ref{tab:main-example} gives example on the three types of questions in two different scenarios, including the question prompt and expected answer.

At test time, to study internal reasoning, a context promp (see appendix) is concatenated before the question to encourage direct answer generation without intermediate reasoning. Greedy decoding ensures deterministic and reproducible outputs, as well as minimizing stochastic noise.

\newcommand{\maxlayer}{footnote for the max layer}

\begin{table*}[ht]
\centering
\begin{adjustbox}{width=\textwidth}
\begin{tabular}{cccccccc}
\toprule
& \multicolumn{7}{c}{Accuracies on MQuake 2-hop Questions (2HQ) Answering} \\
\cmidrule(lr){2-8}
\multirow{2}{*}{Layer(s) to edit} & \multicolumn{3}{c}{Edit Hop-1} & \multicolumn{3}{c}{Edit Hop-2} &\multirow{2}{*}{\textbf{Ave. 2HQ}}  \\
% \cmidrule(lr){2-5} \cmidrule(lr){6-9}
%  & \multicolumn{2}{c}{ROME success} & \multicolumn{2}{c}{2-hop MHQ acc} & \multicolumn{2}{c}{ROME success} & \multicolumn{2}{c}{2-hop MHQ acc} \\
\cmidrule(lr){2-4} \cmidrule(lr){5-7}
 & Hop1(gen.) & Hop2(spec.) & \textbf{2HQ}  & Hop1(spec.) & Hop2(gen.) & \textbf{2HQ} \\
\midrule
5  & 92.5 & 90.8 & \textbf{28.3}  & 74.9 & 79.7 & 3.9& 16.1 \\
10  & 89.6 & 90.8 & 22.5  & 77.2 & 72.1 & 6.7 & 14.6 \\
15  & 72.5 & 90.4 & 14.2  & \textbf{79.7} & 52.6 & 8.9& 11.6 \\
20  & 31.7 & \textbf{91.2} & 3.3  & 77.4 & 28.1 & 10.0 & 6.7\\
5,15  & 95.8 & 90.4 & 27.1  & 52.6  & 59.6 & 7.5  & 17.3\\
5,20  & 95.4 & 90.0 & 27.5  & 53.5 & 60.7 & 6.4 & 17.0\\
5,10,20  & 96.7 & 90.4 & 27.9  & 70.9 & 88.5 & 12.5 & 20.2\\
5,10,15,20 & 96.7 & 90.4 & 25.4 & 67.1 & 88.9 & 16.2 & 20.4\\
5,9,13,17,20 & \textbf{97.5} & 90.8 & 22.5 & 62.7 & 89.4 & 25.3 & 23.9\\
5,8,11,15,17,20 & \textbf{97.5} & 88.3 & 23.3 & 50.1 & \textbf{91.6} & \textbf{39.8} & \textbf{31.6}\\

\bottomrule
\end{tabular}
\end{adjustbox}
\caption{Accuracies for different edition configurations on MQuAKE 2HQ with single-hop edits. We stop at redundant-editing 6 layers since it starts to show clear failures in COUNTERFACT language metrics (Table~\ref{results-cf} and Figure~\ref{fig:trade-off})}.
\label{tab:main}
\end{table*}
% \input{tables/main-tab-example}

% \subsection{Experiment on Knowledge Editing Performance using R-ROME}

% In this subsection, we present a comprehensive evaluation of the upgraded R-ROME method, a complete overhaul of the original ROME approach. R-ROME addresses one of the most important limitations inherent in ROME---namely, the occasional editing failure where updates fail to propagate effectively through the model. This improvement motivated us to conduct a full-scale experiment to assess its performance.

\subsection{COUNTERFACT Experiment Setup}
This experiment was conducted using the GPT-J-6B model. Our evaluation focused on testing all combinations of editing layers ranging from 5th to 20th, with both vanilla ROME and Redundant editing strategies. 
% We selected these particular layers since limiting the number of layers helps reduce processing time and resource demands. By focusing on mid-range layers, we aim to balance effective knowledge updates with minimal disruption to overall model performance.
We evaluated the edited model using 100 instances from the COUNTERFACT data from \citet{meng2023rome}. For ground truth $(s,r,o^c)$, false facts $(s,r,o^*)$, we measure:
% We evaluated the edited model using the COUNTERFACT benchmark from \citet{meng2023rome}, selecting the first 100 instances (can be considered as random based on how this benchmark dataset is constructed). For ground truth $(s,r,o^c)$, false facts $(s,r,o^*)$, we measure:
\begin{itemize}
    \item \textbf{Efficacy:} Quantifies the shift in model probabilities from the target (edited) fact $P(o^*|s,r)$ to the original fact $P(o^c|s,r)$. The \textbf{Efficacy Score (ES)} is the fraction of counterfactual cases for which $P(o^*|s,r) > P(o^c|s,r)$.
    
    \item \textbf{Generalization:} To assess whether the edit generalizes beyond the exact prompt, the updated model is tested on a set of paraphrased prompts that are semantically equivalent to the original factual query $(s,r)$. For each paraphrase, we check if the edited fact is preferred (i.e. $P(o^*|s,r) > P(o^c|s,r)$) in the new context. The \textbf{Paraphrase Score (PS)} is then the fraction of paraphrases for which this holds.
    
    \item \textbf{Specificity:} Ensures edits do not affect unrelated facts. Evaluated using neighboring subjects $s_n$ satisfying $(s_n, r, o^c)$. We require that the model still prefers the original fact (i.e. $P(o_c|s,r) > P(o^*|s,r)$). The \textbf{Neighborhood Score (NS)} is the fraction of such cases.
    
    \item \textbf{Fluency:} This measures the naturalness of the generated text by computing the weighted average of bi- and tri-gram entropies. Specifically, the fluency score is defined as
    \[
    GE = -\sum_{k} f(k) \log_2 f(k),
    \]
    where \(f(k)\) is the frequency distribution over the observed \(n\)-grams (with \(n = 2,3\)) in the generated text. A lower GE indicates a higher degree of repetitiveness, suggesting degraded fluency.
    
    \item \textbf{Consistency:} To measure how well the generated outputs maintain the intended semantic content (i.e., reflect the inserted fact), we compute the unigram TF-IDF vectors for both the generated text and a reference corpus of texts related to the target property \(o^*\). The consistency score is defined as the cosine similarity between these two TF-IDF vectors:
    \[
    RS = \frac{\langle \text{TFIDF}_{\text{gen}}, \text{TFIDF}_{\text{ref}} \rangle}{\|\text{TFIDF}_{\text{gen}}\| \, \|\text{TFIDF}_{\text{ref}}\|}.
    \]
    A higher RS indicates that the generation is semantically coherent with the target property.
    
    \item \textbf{Score:} This is a comprehensive indicator of the overall language capability of the edited model $G^*$, calculated as:
    \[
    S = \operatorname{Avg}\{ES,\, PS,\, NS,\, \frac{GE_{G*}}{GE_{G}}, RS\},
    \]
    where we normalized the fluency score with respect to the baseline flunecy score under the unedited model $G$ to make it consistent to the other metrics(as percentage).

\end{itemize}

\begin{table*}[ht]
\centering
\footnotesize
\setlength{\tabcolsep}{3pt}
\begin{adjustbox}{width=\textwidth}
\begin{tabular}{c c c c c c c c}
\toprule
\multirow{2}{*}{\textbf{Layer(s)}} & \multicolumn{6}{c}{\textbf{COUNTERFACT}} & \multirow{2}{*}{\textbf{MHQ Acc.}}  \\
\cmidrule(lr){2-7} 
 & \textbf{Score} & \textbf{Efficacy} & \textbf{Generalization} & \textbf{Specificity} & \textbf{Fluency} & \textbf{Consistency} \\
\midrule
5 & 90.9 & 100.0 & 99.5 & 76.3 & 620.9 & 78.8 & 16.1 \\
10 & 89.3 & 100.0 & 98.5 & 69.3 & 617.9 & 79.2 & 14.6 \\
15 & 85.2 & 97.0 & 92.5 & 65.6 & 602.1 & 73.9 & 11.6 \\
20 & 76.3 & 94.0 & 73.0 & 65.6 & 543.2 & 61.4 & 6.7 \\
% 5,10 & 86.2 & 100.0 & 100.0 & 60.7 & 604.7 & 73.3 & 24.0 \\
5,15 & 85.4 & 100.0 & 100.0 & 58.8 & 603.9 & 71.0 & 17.3 \\
5,20 & 78.4 & 100.0 & 99.0 & 60.8 & 515.4 & 49.4 & 17.0 \\
% 10,15 & 81.7 & 100.0 & 100.0 & 51.0 & 575.1 & 64.8 & 25.0 \\
% 10,20 & 77.1 & 100.0 & 100.0 & 56.0 & 504.7 & 48.4 & 33.0 \\
% 15,20 & 72.2 & 100.0 & 97.5 & 52.0 & 437.5 & 40.9 & 18.0 \\
% 5,10,15 & 79.2 & 100.0 & 100.0 & 46.7 & 560.1 & 59.2 & 16.0 \\
5,10,20 & 73.8 & 100.0 & 100.0 & 50.8 & 461.8 & 44.1 & 20.2 \\
% 5,15,20 & 71.7 & 100.0 & 100.0 & 47.2 & 440.2 & 40.4 & 22.0 \\
% 10,15,20 & 68.5 & 100.0 & 100.0 & 38.8 & 419.3 & 36.5 & 40.0 \\
5,10,15,20 & 67.1 & 100.0 & 100.0 & 38.8 & 387.7 & 34.2 & 20.4 \\
5,9,13,17,20 & 54.9 & 100.0 & 99.5 & 16.8 & 255.2 & 17.1 & 23.9 \\
5,8,11,15,17,20 & 56.6 & 100.0 & 99.5 & 17.0 & 289.8 & 19.7 & 31.6 \\
\bottomrule
\end{tabular}
\end{adjustbox}
\caption{COUNTERFACT experiment results, alongside the respective MQuAKE multi-hop question answering accuracy for each layer combination, for all metrics, larger the better. We stop at redundant-editing 6 layers, since it starts to show clear failures in the score.}
\label{tab:cf}
\end{table*}

% \input{tables/trade-off-plot}

% \begin{figure*}[ht]
% \centering
% \includegraphics[width=\linewidth]{images/Combined_TOP_SCORE.png}
% \caption{Trade-off between Top 1 and Top 3 Accuracy versus Score.}
% \label{fig:fig1}
% \end{figure*}

\section{Results and Discussion}

Our experiments reveal a clear trade-off in model editing performance: while the Redundant Editing strategy significantly enhances multi-hop reasoning capabilities as evaluated on the MQuAKE dataset, it concurrently results in poorer naturalness metrics on single-hop reasoning tasks, exemplified by performance on the COUNTERFACT dataset. 
We analyze these effects separately in Sections~\ref{results-mq} and~\ref{results-cf}, followed by a comprehensive trade-off analysis illustrated in Figure~\ref{fig:trade-off}. 
Given these insights, practitioners are encouraged to select editing strategies aligned with their specific task objectives: prioritizing multi-hop reasoning for compositional tasks or single-hop naturalness for simpler, fact-based applications.

\subsection{MQuAKE Results Evaluation}
\label{results-mq}

Table~\ref{tab:main} presents the accuracies for various layer editing configurations on the 2HQ in MQuAKE. For single-layer edits, the results align with out hypothesis about knowledge storage: early layer (e.g., layer 5) excels hop-1 reasoning accuracy at 28.3\%, compared to late layers (e.g., layer 20) at 3.3\%. On the other hand, late layers perform better at hop-2 edits with an accuracy of 10.0\% for layer 20 and  3.9\% for layer 5. Additionally, we observe a decreasing trend for generalization ability in both scenarios (edit hop-1 and hop-2) on single-hop questions as deeper layers are involved.

Employing a Redundant Editing approach substantially improves the model's capability to handle multi-hop reasoning tasks.
Editing layers \textbf{5, 8, 11, 15, 17, and 20} achieves the highest average two-hop reasoning accuracy of \textbf{31.6\%}, demonstrating significant improvement over configurations involving fewer layers (e.g., single-layer edit at layer 5 yield only 16.1\% accuracy). This improvement comes from (1) Redundant Editing improves the generalisability of edited knowledge, makes it queriable under different rephrasing of the prompt question.  (2) Redundant Editing creates more possible internal reasoning chains. More comprehensive examinations of these phenomena appear in sections~\ref{analysis:generalisation} and ~\ref{analysis:hopping}.

\subsection{COUNTERFACT Results Evaluation}
\label{results-cf}
Table~\ref{tab:cf} presents the results from the COUNTERFACT dataset, highlighting a notable decreasing trend in naturalness metrics as number of layers edited increases. 
% While single-layer edits, particularly at layer 5, yield high scores in \textbf{specificity} (76.3) and \textbf{fluency} (620.9)
% , these metrics consistently deteriorate when we redundantly edit additional layers.

Specificity decreases significantly from \textbf{76.3} (layer 5 alone) to \textbf{17.0} (layers 5, 8, 11, 15, 17, 20). Similarly, fluency scores decline sharply from \textbf{620.9} (layer 5 alone) to \textbf{255.2} (layers 5, 9, 13, 17, 20) and this trend continues as more layers are involved. This indicates that editing multiple layers simultaneously negatively impacts the coherence and naturalness of single-hop fact recall in the model. We provide a detailed analysis in section~\ref{analysis:overfitting}

The overall COUNTERFACT \textbf{Score} metric also reflects this decreasing trend, declining from \textbf{90.9} for single-layer edits (layer 5) to \textbf{56.6} for Redundant Editing with the 6 layers (layers 5, 8, 11, 15, 17, 20). Thus, these results underscore the trade-off involved in redundancy: while beneficial for multi-hop reasoning, it significantly reduces naturalness and single-hop specificity. Practitioners prioritizing factual naturalness should therefore prefer editing fewer layers, focusing on earlier model layers to maintain optimal single-hop performance.

% For optimal factual naturalness, practitioners should limit edits to fewer, earlier layers.

\section{Failure Patterns of ROME on MQuAKE Questions}

\subsection{ROME Fails in Generalization When Editing Higher Layers}
\label{analysis:generalisation}

We observe that while ROME achieves stable and high edit success rates, its generalization to rephrased prompts degrades markably in higher layers. This limitation persists even when knowledge is inserted at the correct hopping position, ultimately failing to produce accurate answers for two-hop questions.

We hypothesize that this generalization gap may stem from the intrinsic mechanism by which ROME updates the weight matrix. In ROME, the weight update is performed via a rank‐one modification of the MLP’s down‐projection matrix at a given layer, and is computed as 
\begin{equation}
  \hat{W} = W + \Lambda (C^{-1} k^*)^\top.
\end{equation}
The key representation, $k^*$, is derived from the activations corresponding to the subject token at the critical final token position. More concretely, $k^*$ is obtained by applying a non-linear transformation to the pre-activation of the MLP at that token, often expressed as 
\begin{equation}
k^* = \sigma\Bigl(W^{(l)}_{fc}\gamma\bigl(a^{(l)} + h^{(l-1)}\bigr)\Bigr),
\end{equation}
where $W^{(l)}_{fc}$ is the first-layer weight matrix of the MLP at layer $l$, $\gamma$ denotes a normalizing nonlinearity, and $a^{(l)}$ and $h^{(l-1)}$ represent the attention and previous layer hidden states, respectively.

The matrix $C$ captures the uncentered covariance of key representations, calculated as $C = K K^\top$, with $K$ being a matrix whose columns are key representations aggregated from a representative sample of context.

Finally, $\Lambda$ is computed to satisfy the constraint that the updated weight matrix yields the desired output for the given key.
\begin{equation}
\Lambda = \frac{v^* - W k^*}{(C^{-1} k^*)^\top k^*}.
\end{equation}
This update not only adjusts the weight matrix in the direction necessary to encode the new fact, but also critically depends on the fidelity of the key representation $k^*$. If the key derived from the original prompt diverges significantly from that obtained from a rephrased prompt, the update may misalign with the new representation, thus affecting the generalizability of the edit.

To test this assumption, we experimented with a GPT-J-6B model using the MQuAKE dataset. For each layer from 5 to 25, we extracted the subject key for both the original and the rephrased version of the editing prompt, aggregating data over the first 500 instances. We then calculated the cosine similarity between the key vectors corresponding to the two prompt variations, quantifying the consistency of the subject's representation in different phrasings.

\begin{figure}[ht]
\begin{tikzpicture}[scale=0.6]
  \begin{axis}[
      xlabel={Layer},
      grid=major,
      legend pos=north east,
      width=12cm,
      height=8cm,
  ]
    % Cosine Similarity values multiplied by 100 to show percentage
    \addplot[mark=o,blue,thick] coordinates {
      (5,78.13208103179932)
      (6,79.19467091560364)
      (7,77.92861461639404)
      (8,76.13912224769592)
      (9,76.79862976074219)
      (10,76.42148733139038)
      (11,77.41517424583435)
      (12,73.58874082565308)
      (13,75.91248750686646)
      (14,71.47707343101501)
      (15,71.64274454116821)
      (16,65.66541790962219)
      (17,64.67142701148987)
      (18,62.82800436019897)
      (19,57.57814049720764)
      (20,54.87348437309265)
    };
    \addlegendentry{Cosine Similarity (\%)}
    
    % Generalization Accuracy data
    \addplot[mark=square*,red,thick] coordinates {
      (5,86.1)
      (8,79.95)
      (10,80.85)
      (13,76.1)
      (15,62.55)
      (18,47.25)
      (20,29.9)
    };
    \addlegendentry{Generalization Accuracy (\%)}
  \end{axis}
\end{tikzpicture}
\caption{Cosine similarity between subject keys extracted from original and rephrased prompts versus layer (blue) and generalization accuracy from MQuAKE versus layer of the edit (red).}
\label{fig:cos_sim}
\end{figure}
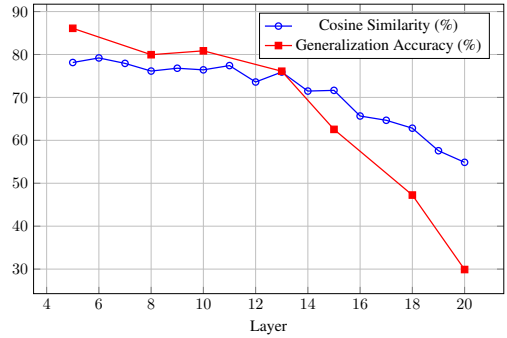

As illustrated in Figure~\ref{fig:cos_sim}, the average cosine similarity between the subject keys for the original and rephrased prompts declines steadily from roughly 0.80 at layer 5 to about 0.50 at layer 25. This downward trend closely parallels the observed drop in generalization performance after the edit, suggesting that increasing divergence in key representations at deeper layers partially drives the degradation. These results highlight the sensitivity of ROME’s rank-one update to variations in the subject’s key and point toward mitigating representational drift as a promising direction for enhancing edit generalizability.

\subsection{Single-Layer ROME Suffers From Hopping-Too-Late}
\label{analysis:hopping}
\begin{figure}[ht]
\centering
\begin{tikzpicture}[scale=0.6]
\begin{axis}[
    xlabel={Layers Edited},
    ylabel={2HQ Accuracy (\%)},
    xmin=5, xmax=20,
    ymin=0, ymax=70,
    xtick={5,8,10,13,15,18,20},
    ytick={0,10,20,30,40,50,60,70},
    legend pos=north east,
    grid=major,
    width=12cm,
    height=8cm,
    ymajorgrids=true,
    xmajorgrids=true,
    tick label style={font=\small},
    label style={font=\small},
    title style={font=\small}
]

% Original Hop1 edited line (light red)
\addplot[red!30, thick, mark=square*] coordinates {
    (5,30.0)
    (8,25.8)
    (10,23.8)
    (13,19.2)
    (15,14.2)
    (18,8.3)
    (20,3.3)
};

% Original Hop2 edited line (light blue)
\addplot[blue!30, thick, mark=triangle*] coordinates {
    (5,3.9)
    (8,7.2)
    (10,6.7)
    (13,10.9)
    (15,8.9)
    (18,8.1)
    (20,10.0)
};

% Normalized Hop1 edited line (standard red)
\addplot[red, thick, mark=square*] coordinates {
    (5,30.0/0.482)
    (8,25.8/0.495)
    (10,23.8/0.482)
    (13,19.2/0.494)
    (15,14.2/0.407)
    (18,8.3/0.341)
    (20,3.3/0.209)
};

% Normalized Hop2 edited line (standard blue)
\addplot[blue, thick, mark=triangle*] coordinates {
    (5,3.9/0.482)
    (8,7.2/0.495)
    (10,6.7/0.482)
    (13,10.9/0.494)
    (15,8.9/0.407)
    (18,8.1/0.341)
    (20,10.0/0.209)
};

\legend{
    Hop1 Edited (raw),
    Hop2 Edited (raw),
    Hop1 Edited (normalized),
    Hop2 Edited (normalized)
}

\end{axis}
\end{tikzpicture}
\caption{2HQ accuracy (raw and generalization-normalized) by edited layer position. Light colors show raw accuracy, while standard colors show accuracy divided by layer-wise generalization accuracy (red points in figure~\ref{fig:cos_sim}), ablating the generalizability decay and studying the underlying editing efficiency independent. 2HQ accuracy by edited layer and hop position, showing inverse patterns for hop-1 (optimal in early layers) and hop-2 (optimal in late layers). Single-layer edits cannot address both requirements simultaneously.}
\label{fig:layer_comparison}
\end{figure}
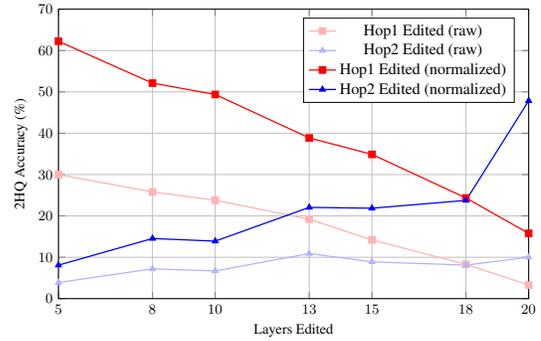
As demonstrated in Figure~\ref{fig:layer_comparison}, the inverse accuracy patterns for hop-1 and hop-2 editing reveal a fundamental limitation of single-layer modifications. The reasoning chain of internal representation dynamics requires hop-1 knowledge to be stored in earlier layers than hop-2 knowledge \citep{DBLP:journals/corr/hopping-too-late}. Since 2HQ reasoning unpredictably uses knowledge as either hop-1 or hop-2 in the test time, editing one single layer forcing an accuracy trade-off between the two scenarios.

Our Redundant Editing strategy overcomes this by simultaneously inserting knowledge copies, for example at layers 5, 8, 11, 15, 17, 20, to ensure optimal positioning for both hops. This approach yields balanced performance when editing hop-1 and hop-2 (Table~\ref{tab:main}) with a 15.5 percentage point (96.3\%) multi-hop accuracy gain compared to the vanilla single edit strategy.

Note that although we can make a minimum of two edits, one at the very early layer and one at the very late layer to ensure the correct hopping order of all the 2-hop questions that require this edited fact, editing later layers causes generalisation decay. Consequently, more layer Redundant Editing achieves higher accuracy because there is a larger chance that a knowledge is in a relative early layer (hence better generalisability) while in the correct hopping order.

\subsection{ROME Overfits a 2HQ to Edited Knowledge When Editing Hop-1}
\label{analysis:overfitting}
% \subsection{Redun}
\begin{table}[ht]
\centering
\begin{tabular}{lccc}
\hline
\textbf{Edited Layers} & \textbf{$|C_{\text{org}}|$} & \textbf{$|C_{\text{abl}}|$} & \textbf{Overfit\%} \\ \hline
% GPT-J (no edit) & 125 & 129 & 3.2 \\
% {[5]} & 121 & 157 & 29.8 \\
% {[10]} & 113 & 157 & 38.9 \\
% {[15]} & 84 & 107 & 27.4 \\
% {[20]} & 30 & 32 & 6.7 \\
% {[5,15]} & 120 & 164 & 36.7 \\
% {[5,20]} & 121 & 160 & 32.2 \\
% {[5,10,20]} & 119 & 162 & 36.1 \\
% {[5,10,15,20]} & 114 & 157 & 37.7 \\ 
% {[5,9,13,17,20]} & 64 & 109 & 41.3 \\ 
% {[5,8,11,15,17,20]} & 23 & 109 & 41.3 \\ 
GPT-J (no edit) & 121 & 125 & 3.2 \\
{[5]} & 85 & 121  & 29.8 \\
{[10]} & 69 & 113 & 38.9 \\
{[15]} & 61 & 84 & 27.4 \\
{[20]} & 28 & 30 & 6.7 \\
{[5,15]} & 76 & 120 & 36.7 \\
{[5,20]} & 82 & 121 & 32.2 \\
{[5,10,20]} & 76 & 119  & 36.1 \\
{[5,10,15,20]} & 71 & 114 & 37.7 \\ 
{[5,9,13,17,20]} & 64 & 109  & 41.3 \\ 
{[5,8,11,15,17,20]} & 23 & 52 & 55.8 \\ 

\hline
\end{tabular}
\caption{Analysis of number of overfitting cases in 2-hop question answering with hop-1 edited by ROME. }
\label{tab:overfitting}
\end{table}
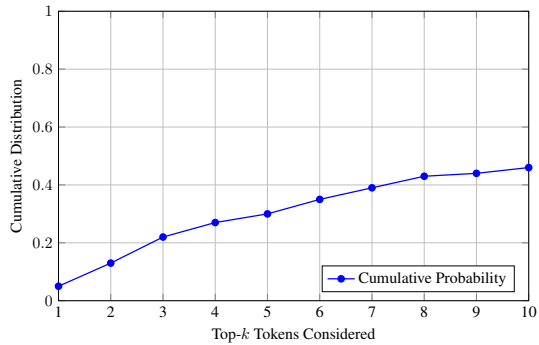
\begin{figure}[ht]
\centering
\begin{tikzpicture}[scale=0.6]
\begin{axis}[
    xlabel={Top-$k$ Tokens Considered},
    ylabel={Cumulative Distribution},
    xmin=1, xmax=10,
    ymin=0, ymax=1,
    xtick={1,2,3,4,5,6,7,8,9,10}, 
    xticklabels={1,2,3,4,5,6,7,8,9,10}, 
    xmode=linear,
    ymajorgrids=true,
    xmajorgrids=true,
    width=12cm,
    height=8cm,
    legend pos=south east
]

% Manual coordinate transformation to compress 10-50k
\addplot[blue, thick, mark=*] coordinates {
    (1, 0.05)
    (2, 0.13) 
    (3, 0.22)
    (4, 0.27)
    (5, 0.3)
    (6, 0.35)
    (7, 0.39)
    (8, 0.43)
    (9, 0.44)
    (10, 0.46)
};

\addlegendentry{Cumulative Probability}

\end{axis}
\end{tikzpicture}
\caption{Post-edit answer accumulative distribution showing persistence of original knowledge: 
in nearly half of cases (46\%), the original answer remains among the top-10 predicted tokens 
even after model editing.}
\label{fig:topk-old-ans}
\end{figure}
In this section, we analyze the \textit{overfitting effect} in 2-hop question answering, where when editing hop-1, models occasionally favor intermediate hop-1 answers over correct final 2HQ answers, even when the correct solution appears high in their predictions. 

As quantified in Table~\ref{tab:overfitting}, we measure this effect through controlled distributional comparisons. Let $C_{\text{origin}}$ denote the cases where the model predicts 2HQ answer correctly, and $C_{\text{ablated}}$ denote the correct cases after hop-1 answer removed from the generation. We have $C_{\text{ablated}} \supseteq C_{\text{origin}}$ (removing interference never reduces correct predictions) and the overfit percentage is computed as: 
\[
\text{Overfit \%} = \left( \frac{|C_{\text{ablated}|} - |C_{\text{origin}}|}{|C_{\text{ablated}}|} \right) \times 100\%
\]

% $(C_{\text{ablated}} - |C_{\text{origin}}|)/C_{\text{ablated}}$.

% As quantified in Table~\ref{tab:overfitting}, we measure this effect through controlled distributional comparisons:

% \begin{enumerate}
%     \item Let $|C_{\text{origin}}|$ denote number of correct cases where the 2HQ answer appears in the top-1 predictions 
%     \item Let $C_{\text{ablated}}$ denote cases where the 2HQ answer ranks top-1 after removing the hop-1 answer. The cases constituting $C_{\text{ablated}}$ form a superset of those in $|C_{\text{origin}}|$, since removing the interfering hop-1 answer can only increase the number of correct top-1 predictions for the 2HQ answer.
% \end{enumerate}

% % The overfitting percentage is then computed as:

This metric captures the relative frequency with which the hop-1 answer incorrectly blocks the 2HQ answer from reaching the top position.
Our experiments compare different layer-editing configurations, revealing that models exhibit significantly higher overfitting (up to $55.8\%$) after ROME edits, whereas the unmodified baseline (GPT-J) shows minimal bias ($3.2\%$). 

The observed overfitting in 2HQ is possibly due to that \textsc{ROME} edits do not erase original knowledge but instead introduce a \textit{stronger competing signal} that dominates the model's outputs. 
This finding is illustrated in Figure~\ref{fig:topk-old-ans}, that the original knowledge often remains accessible in the top-$k$ predictions for edited facts. 
The success of ROME hinges on this signal strength overriding the original association, but it inadvertently disrupts multi-hop reasoning by over-activating intermediate (hop-1) answers at the expense of later-hop deductions. 
This behavior is consistent with the hypothesis that knowledge edits operate via signal interference rather than overwriting old knowledge, as evidenced by the lack of correlation between localized knowledge positions and edit success~\cite{hase2023does}. 
Note that Redundant Editing amplifies this effect. Inserting more knowledge copies further strengthens the dominant signal, which explains its observed trade-off of lower specificity for higher multi-hop accuracy.

\section{Conclusion}
This work addresses critical limitations in knowledge editing for multi-hop reasoning. Through systematic analysis of ROME's failure patterns, including the hopping-too-late problem, generalization decay and overfitting issue. We develop Redundant Editing, which strategically distributes knowledge across multiple network layers. Our approach achieves a 15.5 percentage point (96\%) improvement in 2-hop questions accuracy while maintaining language quality. We also study the trade-off between multi-hop reasoning ability and language metrics, including the specificity and naturalness.

\section{Limitations and Future Work}
Our work has several limitations that suggest productive directions for future research. While we demonstrate the effectiveness of Redundant Editing within the ROME framework, our analysis does not extend to other knowledge editing methods (e.g., fine-tuning or representation editing \citep{hernandez2023inspecting}) or alternative model architectures (e.g., encoder-decoder or sparse models). Additionally, our experiments are confined to 2-hop questions with single-hop edits, leaving open questions about the scalability to ($n \geq 3$)-hop reasoning and the effects of simultaneously editing multiple hops. These unexplored dimensions represent important avenues for future advancements in knowledge editing research.

% \clear page
\bibliography{references}

% \iftaclpubformat

\onecolumn

\appendix
\section{2-Hop Question Prompt Contexts}
\label{app:prompts}
\begin{table*}[ht]
\centering
\begin{tabular}{p{\textwidth}}
\toprule
"context": \\ "Q: What is the name of the current head of state in Newfoundland and Labrador? A: Elizabeth II\\Q: What is the name of the current head of state in United States of America? A: Donald Trump\\Q: What is the name of the current head of state in Stoltenberg's Second Cabinet? A: Harald V of Norway\\Q: What is the name of the current head of state in Germany? A: Frank-Walter Steinmeier\\Q: What is the name of the current head of state in India? A: Ram Nath Kovind\\Q: What is the name of the current head of state in Manipur? A: Najma Heptulla\\Q: What is the name of the current head of state in France? A: Emmanuel Macron\\Q: What is the name of the current head of state in Uttarakhand? A: Krishan Kant Paul"\\
\midrule
"question": "Q: What is the name of the current head of state in the United Kingdom? A:"\\
\bottomrule
\end{tabular}
\caption{Context prompt for the question to encourage the model generate answers directly without thinking process.}
\label{app-prompt-context}
\end{table*}

\end{document}